\documentclass[letterpaper, 10 pt, conference]{ieeeconf}  % Comment this line out if you need a4paper

\IEEEoverridecommandlockouts                              % This command is only needed if 
\usepackage{graphicx}
\usepackage{algorithm}
\usepackage{algorithmic}
\usepackage{comment}
\usepackage{multirow}
\usepackage{booktabs} 
\usepackage{cite}
\title{\LARGE \bf
REFLEX: Metacognitive Reasoning for Reflective Zero-Shot Robotic Planning with Large Language Models
}

\author{Wenjie Lin$^{1}$, Jin Wei-Kocsis$^{1*}$, Jiansong Zhang$^{2}$, Byung-Cheol Min$^{1}$, Dongming Gan$^{3}$,\\ Paul Asunda$^{4}$, and Ragu Athinarayanan$^{3}$%
\thanks{$^{*}$Corresponding author.}%
\thanks{$^{1}$Wenjie Lin and Jin Wei-Kocsis are with the School of Applied and Creative Computing, Purdue University, West Lafayette, IN, USA; Byung-Cheol Min was with the School of Applied and Creative Computing, Purdue University, West Lafayette, IN, USA 
        {\tt\small \{lin1790,kocsis0,minb\}@purdue.edu}}%
\thanks{$^{2}$Jiansong Zhang is with the Bowen School of Construction, Purdue University, USA 
        {\tt\small zhan3062@purdue.edu}}%
\thanks{$^{3}$Dongming Gan and Ragu Athinarayanan are with the School of Engineering Technology, Purdue University, USA
        {\tt\small \{dgan,rathinar\}@purdue.edu}}%
\thanks{$^{4}$Paul Asunda is with the Department of Technology Leadership and Innovation, Purdue University, USA
        {\tt\small pasunda@purdue.edu}}%
}
\begin{document}

\maketitle
\thispagestyle{empty}
\pagestyle{empty}

%%%%%%%%%%%%%%%%%%%%%%%%%%%%%%%%%%%%%%%%%%%%%%%%%%%%%%%%%%%%%%%%%%%%%%%%%%%%%%%%
\begin{abstract}
While large language models (LLMs) have shown great potential across various domains, their applications in robotics remain largely limited to static prompt-based behaviors and still face challenges in complex tasks under zero-shot or few-shot settings. Inspired by human metacognitive learning and creative problem-solving, we address this limitation by exploring a fundamental question: \emph{Can LLMs be empowered with metacognitive capabilities to reason, reflect, and create, thereby enhancing their ability to perform robotic tasks with minimal demonstrations?} In this paper, we present REFLEX, a framework that integrates metacognitive learning into LLM-powered multi-robot collaboration. The system equips the LLM-powered robotic agents with a skill decomposition and self-reflection mechanism that identifies modular skills from prior tasks, reflects on failures in unseen task scenarios, and synthesizes effective new solutions. We propose a more challenging robotic benchmark task and evaluate our framework on the existing benchmark and the novel task. Experimental results show that our metacognitive learning framework significantly outperforms existing baselines. Moreover, we observe that our framework can generate solutions that differ from the ground truth yet still successfully complete the tasks. These findings support our hypothesis that metacognitive learning can foster creativity in robotic planning.
\end{abstract}

\section{Introduction}\label{Introduction} 

\begin{figure*}[ht]
  \includegraphics[width=\linewidth]{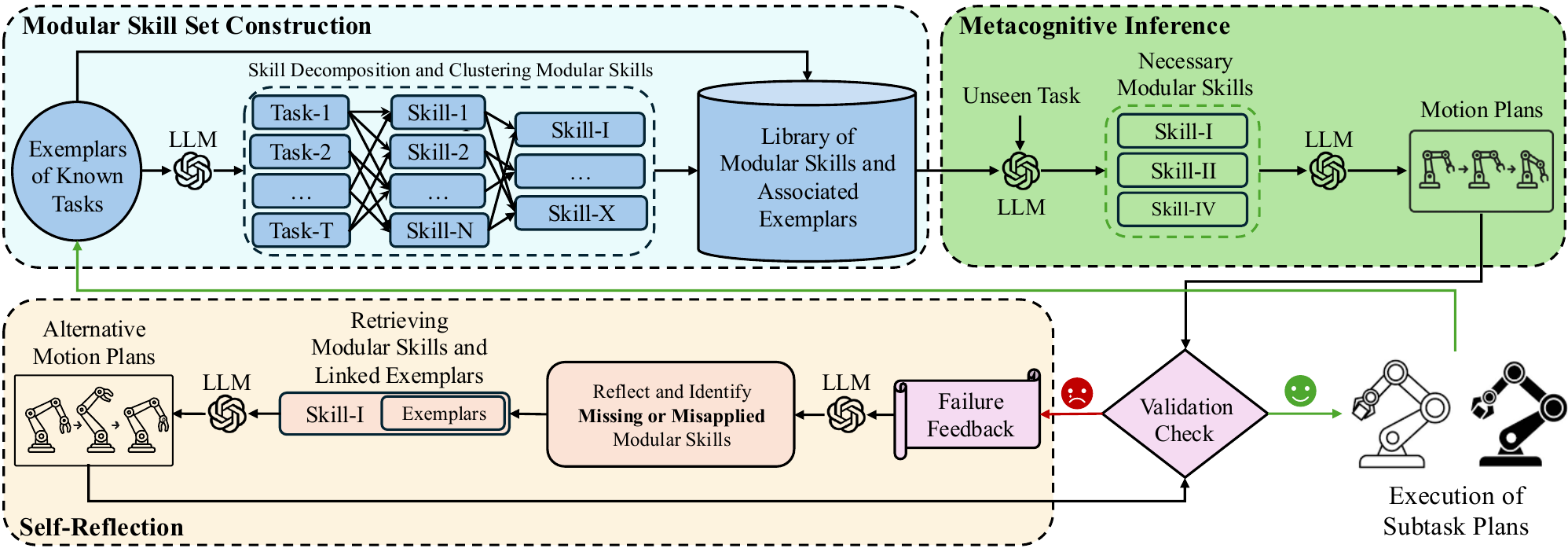}
  \caption {Overview of the proposed REFLEX framework. The framework consists of three interconnected components: (1) modular skill set construction, where prior successful task exemplars are decomposed and clustered into reusable modular manipulation skills; (2) metacognitive inference, where relevant skills and associated exemplars are selectively retrieved and composed to synthesize motion plans for unseen tasks; and (3) structured self-reflection, where validation feedback updates the metacognition-informed input to diagnose missing or misapplied skills and regenerate improved plans. This closed-loop reasoning mechanism enables reliable recovery and adaptive alternative plan generation under zero-shot multi-robot collaboration settings.}
  \label{fig:overview}
\end{figure*}
In recent years, LLMs have emerged as powerful reasoning engines capable of performing complex planning, decision-making, and knowledge-intensive tasks across various application domains. These successes have sparked growing interest in applying LLMs to robotic operations, enabling robots to understand instructions, generate executable action sequences, and generalize across diverse and novel scenarios~\cite{wang2024large, liu2024enhancing, jin2024robotgpt, tan2024multi, cheng2024empowering, chen2025robohorizon}. 

Recent research on LLM-powered robotic operations can be broadly categorized into three main directions: 1) generating robot plans by prompting LLMs with task instructions~\cite{mandi2024roco}, 2) enabling embodied reasoning by integrating LLMs with multimodal perception systems~\cite{aissi2025viper}, and 3) synthesizing robot control codes from natural language commands~\cite{liang2023code}. Within these directions, structured planning frameworks such as AutoTAMP~\cite{chen2024autotamp} and Symbolic Planner~\cite{chen2024language} combine language models with hybrid or symbolic planning mechanisms for task decomposition and execution. In parallel, systems such as SayPlan~\cite{rana2023sayplan}, VoxPoser~\cite{huang2023voxposer}, Code as Policies~\cite{liang2023code}, and RoCo~\cite{mandi2024roco}, study language-guided planning for spatial grounding, interpretable program generation, and multi-robot collaboration. While these approaches demonstrate the potential of LLMs to advance robotic operations, most existing approaches lack mechanisms for metacognitive reasoning or dynamic adaptation after failures, restricting their effectiveness in complex zero-shot or few-shot tasks.

Inspired by human metacognitive learning and its impact on enabling creative skills~\cite{hargrove2013assessing, schuster2020transfer}, we address this limitation by exploring a fundamental question: \emph{Can LLMs be endowed with metacognitive capabilities to reason, reflect, and create, thereby enhancing their ability to perform robotic tasks with minimal demonstrations?} In this paper, we present an early-stage framework that integrates metacognitive learning into LLM-powered multi-robot collaboration. The proposed system equips the LLM with skill decomposition, metacognitive inference, and self-reflection mechanisms that identify modular skills from prior tasks, reflect on failures in novel scenarios, and synthesize effective new solutions. Related large-scale embodied systems, such as RT-1/RT-X~\cite{brohan2022rt,o2024open}, OpenVLA~\cite{kim2024openvla}, Octo~\cite{team2024octo}, RoboBallet~\cite{lai2025roboballet}, and RoboGen~\cite{wang2023robogen}, study robot behavior through large-scale robot data, vision-language-action modeling, reinforcement learning, graph-based reasoning, or simulation-driven training. In contrast, our focus is on text-based metacognitive reasoning for multi-robot planning. Our work makes three key contributions:
\begin{itemize}
    \item To the best of our knowledge, this is the first work to explore integrating metacognitive learning into LLM-equipped robot manipulation, to support both reliable performance and creative problem-solving; 
    \item  We propose REFLEX, a metacognitive learning framework that enables the LLM-powered robotic agents decompose modular skills, metacognitive inference, reflect on task failures, and synthesize effective new solutions;
    \item We develop a novel robotic benchmark task and validate our framework on both our new and existing benchmark tasks, where it significantly outperforms baselines and sometimes generates valid solutions that deviate from the ground truth, which supports the hypothesis that metacognitive learning can foster reliable and creative robotic planning. 
\end{itemize}

The rest of the paper is organized as follows. We first present related work and problem formulation, and describe our metacognitive learning framework. Experiments and results follow, along with discussions and limitations. We conclude in the final section.

\section{Related Work}
\label{Related Work}

This section reviews prior work most relevant to our study from three perspectives: LLM-based robotic systems, metacognitive learning, and creativity in robotic planning. Together, these areas provide the foundation for our metacognitive framework REFLEX for reflective multi-robot planning.

\paragraph{LLM-powered Robotic Systems} LLMs have demonstrated great potential in robotic planning and reasoning. Recent advances include prompt-based planning~\cite{mandi2024roco}, multimodal perception integration~\cite{tan2024multi,aissi2025viper}, and direct synthesis of robot control code from natural language~\cite{liang2023code,chen2025robohorizon}. SayPlan~\cite{rana2023sayplan} studies language-guided planning for robotic task execution. VoxPoser~\cite{huang2023voxposer} investigates language-guided planning with spatial grounding through perception and affordance-based control. Code as Policies~\cite{liang2023code} focuses on interpretable program generation from natural language instructions. RoCo~\cite{mandi2024roco} proposed a dialectic approach for LLM-powered multi-robot collaboration. VIPER \cite{aissi2025viper} was introduced by integrating LLMs with visual inputs for sequential decision-making, but without reflection or creative capabilities. Researchers addressed long-horizon planning through multimodal skill learning~\cite{tan2024multi} and multi-view world model~\cite{chen2025robohorizon}, respectively, yet neither incorporated metacognitive reasoning. Direct policy generation was enabled from natural language~\cite{liang2023code} but relied on predefined tasks without adaptive reflection.

\paragraph{Metacognitive Learning} Metacognitive learning, focused on regulation of cognitive processes, has proven beneficial to enhance creativity and problem-solving skills in humans~\cite{hargrove2013assessing,schuster2020transfer}, but its application in robotics remains limited. Prior studies highlight its benefits for learning efficiency, adaptability, and creative skill development in education contexts~\cite{hargrove2013assessing,schuster2020transfer}, providing a foundation for our work. Although some efforts in LLMs and robotics explore self-reflection using reinforcement or iterative learning, they often lack explicit integration of metacognitive reasoning. Our work addresses this gap by explicitly embedding metacognitive components into LLM-driven robotic collaboration.

%\subsection*{Creativity in Robotic Planning}

\paragraph{Creativity in Robotic Planning}
Creativity in robotic task planning often implies the ability to generate effective solutions beyond explicit training data or demonstrations. Prior research on creativity has largely focused on writing or art, emphasizing adaptive generalization or task reformation~\cite{Gómez-Rodríguez2025}. To the best of our knowledge, no existing work focuses on creativity in robot manipulation, where solutions must operate under complex physical and operational constraints. We take a step toward exploring the capabilities of a metacognition learning-based framework to enable structured creativity in robotic planning. In this paper, structured creativity is defined as generating operationally distinct but valid plans that achieve success. Our work significantly expands the creative potential of robotic systems in zero-shot and minimal-demonstration contexts.

\section{Problem Formulation}\label{Problem Formulation}

\begin{figure*}[ht]
  \includegraphics[width=\linewidth]{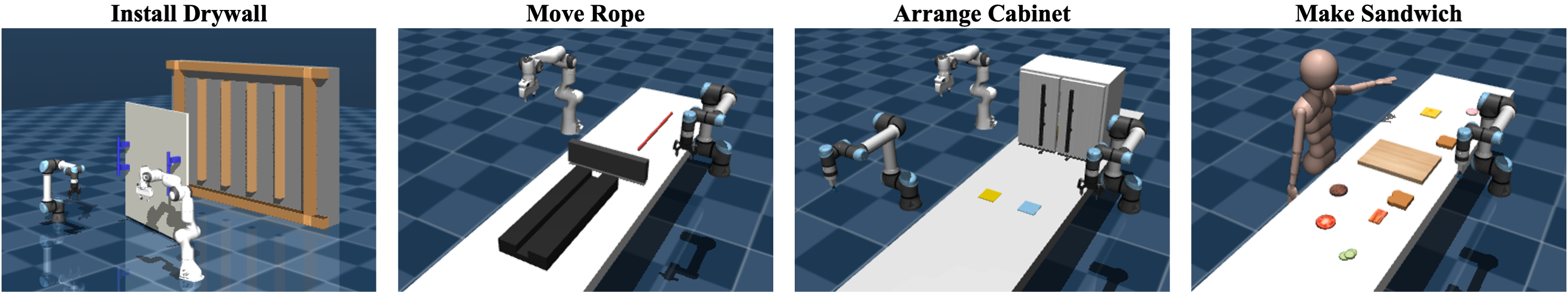}
  \caption {We create a new benchmark robotic task: Install Drywall. Together with the three most challenging tasks in RoCoBench \cite{mandi2024roco}, we test and compare our metacognitive learning framework on the four tasks with the baseline methods.}
  \label{fig:teaser}
\end{figure*}

In our work, we consider a cooperative multi-agent manipulation setting, where multiple LLM-powered robot agents collaborate to complete tasks over a finite time horizon. Each agent operates within its own observation space and needs to coordinate with other agents to achieve a shared task objective. At each time step, each agent~$n$ receives a prompt $p_t^n = f_n(g_n,o_t^n,r_t^n)$ and outputs $\pi_n$, where $\pi_n$ denotes the resulting arm motion plan, $g_n$ represents the agent-specific task description that includes goals and constraints, $o_t^n \in \Omega_n$ is its current observation, and $r_t^n$ is the metacognition-informed input. 

The metacognition-informed input $r_t^n$ is a dynamic guidance signal embedded in the structured prompt provided to each LLM-powered robot agent. It evolves across the three components of the metacognitive learning module and provides essential context that enables the LLM to (1) decompose prior task completions into modular skills, (2) synthesize arm motion plans $\left\{\pi_n\right\}_{n=1}^N$ for unseen task scenarios, (3) reflect on planning generation failures, and (4) iteratively produce effective and potentially creative solutions. More details on the role and structure of $r_t^n$ at each metacognitive-learning stage will be introduced in the following section.

\section{Methodology}
\label{Methodology}

The proposed metacognitive learning framework REFLEX is illustrated in Fig.~\ref{fig:overview}. It comprises three key components: (1) modular skill set construction, (2) metacognitive inference, and (3) self-reflection.

In the \textit{modular skill set construction} component, the LLM is guided by the metacognition-informed input $r_t^n$, which includes retrieved exemplars from previously successful tasks and structured prompt content designed to guide the LLM to support skill abstraction and cluster modular skills. Each exemplar contains a brief task summary, a representative scene, and a one-shot demonstration of an effective action plan. Guided by $r_t^n$, the LLM then extracts robot manipulation skills from each exemplar, clusters similar skills that correspond to similar goals to reduce redundancy, and organizes them into a reusable library of transferable modular robot manipulation skills and their associated exemplars.

In the \textit{metacognitive inference} component, the LLM receives the task description $g_n$, the current observation $o_t^n$, and access to the previously constructed library of modular skills and associated exemplars. The metacognition-informed input $r_t^n$ serves as a reasoning-oriented guidance signal that provides attention cues to focus the LLM's inference. Rather than specifying relevant skills directly, $r_t^n$ helps the LLM reason about which modular skills from the library are applicable to the current task. Using the identified skills and their associated exemplars, the LLM synthesizes arm motion plans $\pi_n$ for the robot agent $n$.

The \textit{self-reflection} component is activated when the arm motion plans synthesized during metacognitive inference fail to pass the validation check (e.g. due to collision or inverse kinematics (IK) infeasibility). In our current implementation, we adopted the validation mechanism in~\cite{mandi2024roco} to detect planning generation failures and trigger the self-reflection process. In this stage, the metacognition-informed input $r_t^n$ is updated to encode structured failure feedback, including the nature and location of the failure. This updated $r_t^n$ serves as a reflective prompt that guides the LLM to reason about which modular skills may be missing or misapplied. Guided by these insights, the LLM retrieves relevant exemplars from the skill library and synthesizes a revised arm motion plan to correct the previous failure.

Equipped with the proposed metacognitive learning module, the LLM adaptively generates reliable and potentially creative arm motion plans, which are subsequently used to update the library of prior task exemplars.

\section{Experiments}\label{Experiments}

% We validate the performance of our proposed metacognitive learning module on empowering LLM-powered robotic agents in completing complex multi-robot collaboration tasks in zero-shot settings. To achieve this, in this initial stage of work, we leverage the RoCo benchmark~\citep{mandi2024roco} for conducting the experiments and target at the four most challenging tasks: Pack Grocery, Arrange Cabinet, Move Rope, and Make Sandwich. In each task, the LLM-powered robot agents coordinate with textual prompts including metacognition input and action execution in environments with obstacles and spatial constraints. The embedded validation mechanism detects planning generation failures, such as the ones caused by collisions and inverse kinematics (IK) feasibility, and triggers the self-reflection process. All experiments are conducted on an NVIDIA A100 GPU to support efficient inference using the LLaMA 3.2-90B Vision model. We would like to clarify that the reason of adopting LLaMA model instead of GPT4 is because that our final objective is to develop a cost-effective open-source LLMs that can be easily adopted by different communities.

We evaluate the performance of the proposed REFLEX framework in enabling LLM-enabled robotic agents to complete complex multi-robot collaboration tasks under zero-shot settings. We conduct experiments using the RoCoBench~\cite{mandi2024roco}, focusing on four challenging tasks: \textit{Move Rope}, \textit{Arrange Cabinet}, \textit{Make Sandwich}, as well as our newly introduced \textit{Install Drywall}. Our \textit{Install Drywall} task represents a significant advancement in multi-robot coordination complexity, requiring continuous safety monitoring, precise spatial alignment, and coordinated load-bearing throughout the task execution process. In all of these tasks, LLM-powered robot agents coordinate via structured textual prompts that include metacognition-informed inputs and execute actions in environments with obstacles and operational constraints. An embedded validation mechanism detects planning failures, such as those caused by collisions or IK infeasibility, and triggers the self-reflection process. All experiments are conducted on a machine with an NVIDIA A100 GPU to support efficient inference using the LLaMA-3.1-70B model and the GPT-4 model for fair comparison.

\begin{table*}[ht]
\caption{Performance comparison between our REFLEX framework and baselines on RoCoBench.}
\centering
\begin{tabular}{ccccc}
\toprule
 & Metrics & \textbf{Move Rope} & \textbf{Arrange Cabinet} & \textbf{Make Sandwich} \\
\midrule
\multirow{2}{*}{Central Plan} & Task Success Rate &
0.50 $\pm$ 0.11 & 
0.90 $\pm$ 0.07 &  
0.96 $\pm$ 0.04 
 \\
 & Environment Steps, Replan Attempts &
2.3, 3.9 & 
4.0, 2.7 &  
8.8, 1.2 \\
\midrule
\multirow{2}{*}{RoCo + GPT-4} & Task Success Rate &
0.65 $\pm$ 0.11 & 
0.75 $\pm$ 0.10 & 
0.80 $\pm$ 0.08 
 \\
 & Environment Steps, Replan Attempts &
2.5, 3.1 & 
4.7, 2.0 & 
10.2, 1.7\\
\midrule
\textbf{REFLEX} + LLaMA-3.1 & Task Success Rate &
0.76 $\pm$ 0.10 & 
0.95 $\pm$ 0.05 & 
0.95 $\pm$ 0.05 
 \\
\textbf{Ours} & Environment Steps, Replan Attempts &
2.0, 2.4 & 
4.0, 1.7 & 
9.4, 1.8 \\
\midrule 
\textbf{REFLEX} + GPT-4 & Task Success Rate &
0.86 $\pm$ 0.08 & 
1.00 $\pm$ 0.00 & 
0.95 $\pm$ 0.05
 \\
\textbf{Ours}  & Environment Steps, Replan Attempts &
2.1, 1.2 & 
4.0, 0.0 & 
9.1, 0.0  \\
\bottomrule
\end{tabular}
\label{tab:rocobench_results}
\end{table*}

\subsection{Benchmark}

The visualization of each task in this paper is shown in Fig.~\ref{fig:teaser}.

\subsubsection{Install Drywall: A New Multi-Robot Construction Task} To evaluate the metacognitive capabilities of LLM-powered robotic systems in complex scenarios resembling real-world jobsites, we design a novel task for multi-robot collaboration, \textit{Install Drywall}. In this task, the robot agents must coordinate to pick up and hold the left and right panel handles, lift the panel to a specified height, detect and correct misalignment between the panel and studs on the wall by applying a twist operation, and finally position the panel against the wall with precise parallel alignment. This task elevates the complexity of multi-robot coordination by explicitly requiring advanced and synchronized spatial reasoning and coordinated physical manipulation. Specifically, two robotic agents must collaborate to install a drywall panel onto a wall frame. Specifically, this task incorporates five distinct modular skills: (1) coordinated dual-agent execution, (2) path and collision planning, (3) spatial reasoning and alignment, (4) conditional or context-aware execution, and (5) object manipulation and transfer. Additionally, the \textit{Install Drywall} task also mandates continuous real-time safety checks, stability monitoring, and precise alignment corrections. Robots must dynamically respond to environmental feedback, reflecting on any encountered failures, and iteratively refining their plans. The task explicitly evaluates the ability of LLM-powered robots to adaptively, and even creatively, generate feasible solutions under challenging zero-shot conditions, thereby effectively validating the robustness and adaptability of our proposed metacognitive learning framework.

\subsubsection{RoCoBench}

The RoCo benchmark \cite{mandi2024roco} is a comprehensive evaluation platform designed explicitly for multi-robot collaboration tasks, addressing a broad spectrum of coordination scenarios. We select the three most challenging and complex tasks from it: \textit{Move Rope}, \textit{Arrange Cabinet}, and \textit{Make Sandwich}. Each task is structured to test specific coordination behaviors, encompassing sequential and parallel task decomposition, varying levels of workspace overlap, and both symmetric and asymmetric observation spaces. The tasks demand sophisticated interaction among robots, including high-level task reasoning, effective information exchange, adaptive sub-task planning, and precise low-level motion planning facilitated by a centralized multi-arm trajectory planner. In our experiment, we leverage the multi-robot collaboration task environment and validation mechanism provided by the RoCoBench, while building our own metacognitive learning-empowered decision-making framework.

%Notably, the benchmark incorporates rigorous validation steps such as collision checking and inverse kinematics feasibility, prompting iterative refinement of robot action plans through structured, dialog-based interactions.

%\subsubsection{A New Task: Install Drywall}

\subsection{Baselines and Performance Metrics}
We compare our framework against two baselines. (1) \textbf{Central Plan}: an oracle LLM-based planner with access to the full environment state, task description, and capabilities of all robots. It generates a joint centralized plan without accounting for information asymmetry. (2) \textbf{RoCo+GPT-4}: the state-of-the-art multi-robot collaboration framework proposed in~\cite{mandi2024roco}, which uses GPT-4 but does not incorporate metacognition-informed input. To ensure a fair performance comparison, we follow RoCoBench~\cite{mandi2024roco} and use the same evaluation metrics: (1) \textit{Task Success Rate}, which measures the percentage of successful task completions within a fixed number of rounds; (2) \textit{Environment Steps}, defined as the average number of steps taken in successful runs; and (3) \textit{Replan Attempts}, which refers to the average number of replan attempts across all runs. For fair comparison with RoCo, we use over 20 rounds, and up to 10 steps and 5 replans per run.

\begin{table}[ht]
\caption{Performance comparison between our REFLEX and baseline on our novel benchmark task. Here, ``success'' represents \textbf{Task Success Rate}; ``step'' means \textbf{Environment Steps}; and ``replan'' refers to \textbf{Replan Attempts}.}
\centering
\begin{tabular}{ccc}
\toprule
& Metrics & \textbf{Install Drywall} \\
\midrule
\multirow{2}{*}{RoCo + GPT-4} & success &
0.62 $\pm$ 0.11
 \\
 & step, replan &
5.3, 5.8 \\
\midrule
\textbf{REFLEX} + LLaMA-3.1 & success &
0.95 $\pm$ 0.05  
 \\
\textbf{(Ours)} & step, replan &
4.2, 1.9  \\
\midrule 
\textbf{REFLEX} + GPT-4 & success&
1.00 $\pm$  0.00
 \\
\textbf{Ours}   & step, replan &
4.1, 0.0 \\
\bottomrule
\end{tabular}
\label{tab:drywall_results}
\end{table}
\section{Results}\label{Results}

\begin{figure*}[ht]
  \centering
  \includegraphics[width=\textwidth]{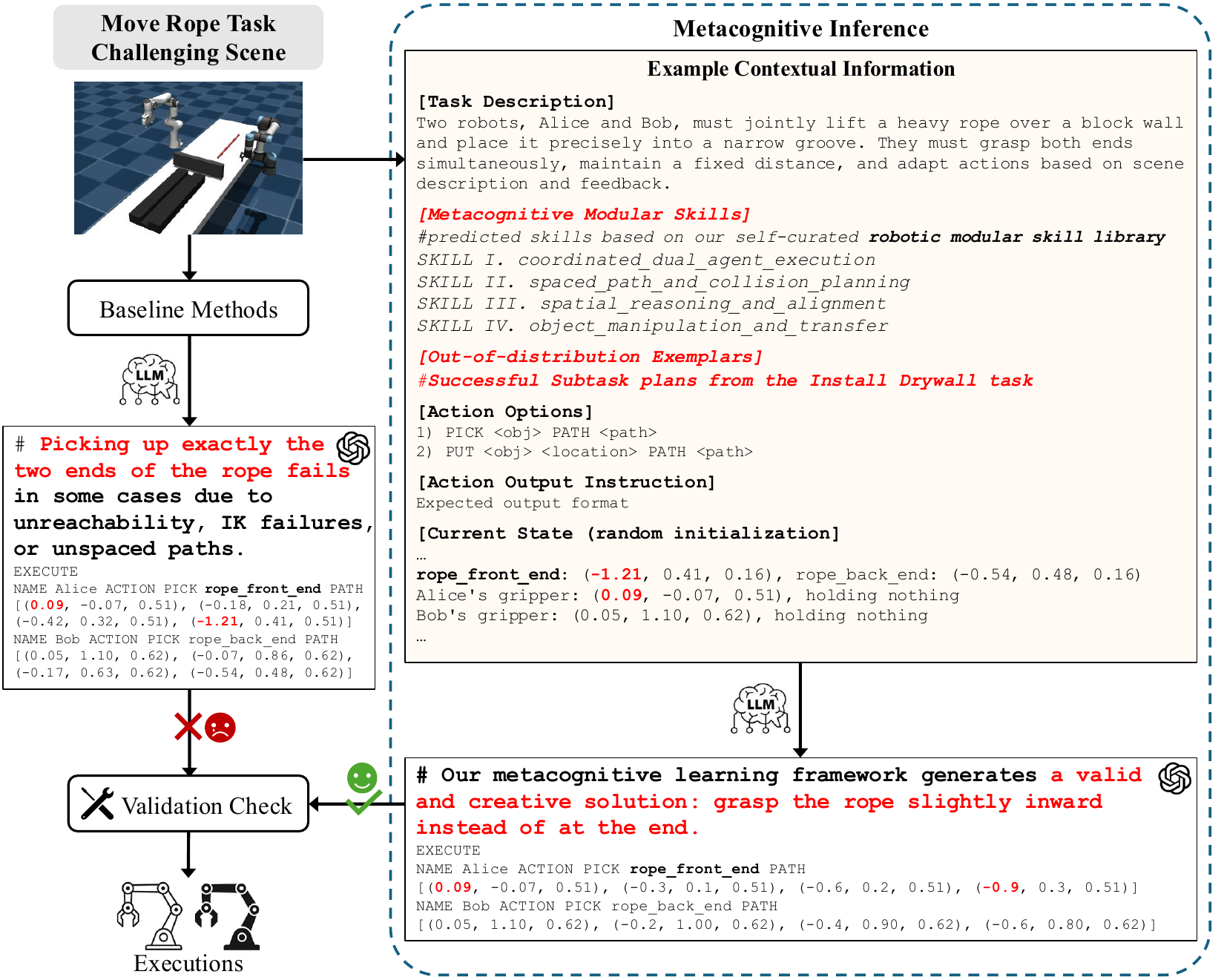}
  \caption{A case study of generating a creative and valid solution by our metacognitive learning framework REFLEX.}
  \label{fig:creativity}
\end{figure*}

\subsection{Modular Robotic Skill Library Construction}

\begin{table*}[ht]
\caption{Skill decomposition across tasks achieved by the proposed REFLEX framework. }
  \centering
  \begin{tabular}{|p{2.7cm}|p{5cm}|p{5.1cm}|}
    \hline
    \textbf{Task} & \textbf{Skills} & \textbf{Necessary Modular Skills} \\
    \hline
    \textbf{Install Drywall} & 
    \begin{tabular}[t]{@{}l@{}}
    synchronized\_dual\_grasping\\
    synchronized\_lifting\\
    pose\_alignment\_checking\\
    panel\_twisting\_and\_adjustment\\
    precise\_object\_positioning
    \end{tabular} &
    \begin{tabular}[t]{@{}l@{}}
    coordinated\_dual\_agent\_execution\\
    spatial\_reasoning\_and\_alignment\\
    object\_manipulation\_and\_transfer\\
    conditional\_or\_contextual\_execution\\
    spaced\_path\_and\_collision\_planning
    \end{tabular} \\
    \hline
    \textbf{Move Rope} & 
    \begin{tabular}[t]{@{}l@{}}
    synchronized\_bimanual\_lifting\\
    obstacle\_avoidance\\
    grasp\_distance\_maintenance
    \end{tabular} &
    \begin{tabular}[t]{@{}l@{}}
    coordinated\_dual\_agent\_execution\\
    spatial\_reasoning\_and\_alignment\\
    spaced\_path\_and\_collision\_planning\\
    object\_manipulation\_and\_transfer
    \end{tabular} \\
    \hline
    \textbf{Arrange Cabinet} & 
    \begin{tabular}[t]{@{}l@{}}
    door\_opening\_mechanics\\
    multi\_arm\_synchronization\\
    gripper\_state\_management\\
    placement\_accuracy
    \end{tabular} &
    \begin{tabular}[t]{@{}l@{}}
    coordinated\_dual\_agent\_execution\\
    spaced\_path\_and\_collision\_planning\\
    object\_manipulation\_and\_transfer\\
    conditional\_or\_contextual\_execution
    \end{tabular} \\
    \hline
    \textbf{Make Sandwich} & 
    \begin{tabular}[t]{@{}l@{}}
    sequential\_object\_stacking\\
    task\_allocation\_under\_constraints\\
    unilateral\_reach\_limitation\_handling\\
    action\_conflict\_avoidance
    \end{tabular} &
    \begin{tabular}[t]{@{}l@{}}
    task\_decomposition\_and\_ordering\\
    spatial\_reasoning\_and\_alignment\\
    conditional\_or\_contextual\_execution\\
    object\_manipulation\_and\_transfer
    \end{tabular} \\
    \hline
  \end{tabular}
  \label{skillset}
\end{table*}

To support generalization across diverse manipulation scenarios, we construct a modular robotic skill library by decomposing complex tasks into reusable modular skills. Our library is grounded in the principle of functional modularity, where each modular skill encapsulates linked exemplar subtask plans and can be recombined across tasks. This enables the LLM-based planner to generalize to novel task scenes and adaptively reconfigure plans in a zero-shot setting.

Table~\ref{skillset} summarizes the mapping from high-level tasks to their corresponding low-level skills and the necessary modular skills, which are achieved by modular skill set construction and metacognitive inference components of our proposed module.

\begin{figure*}[ht]
  \centering
  \includegraphics[width=\textwidth]{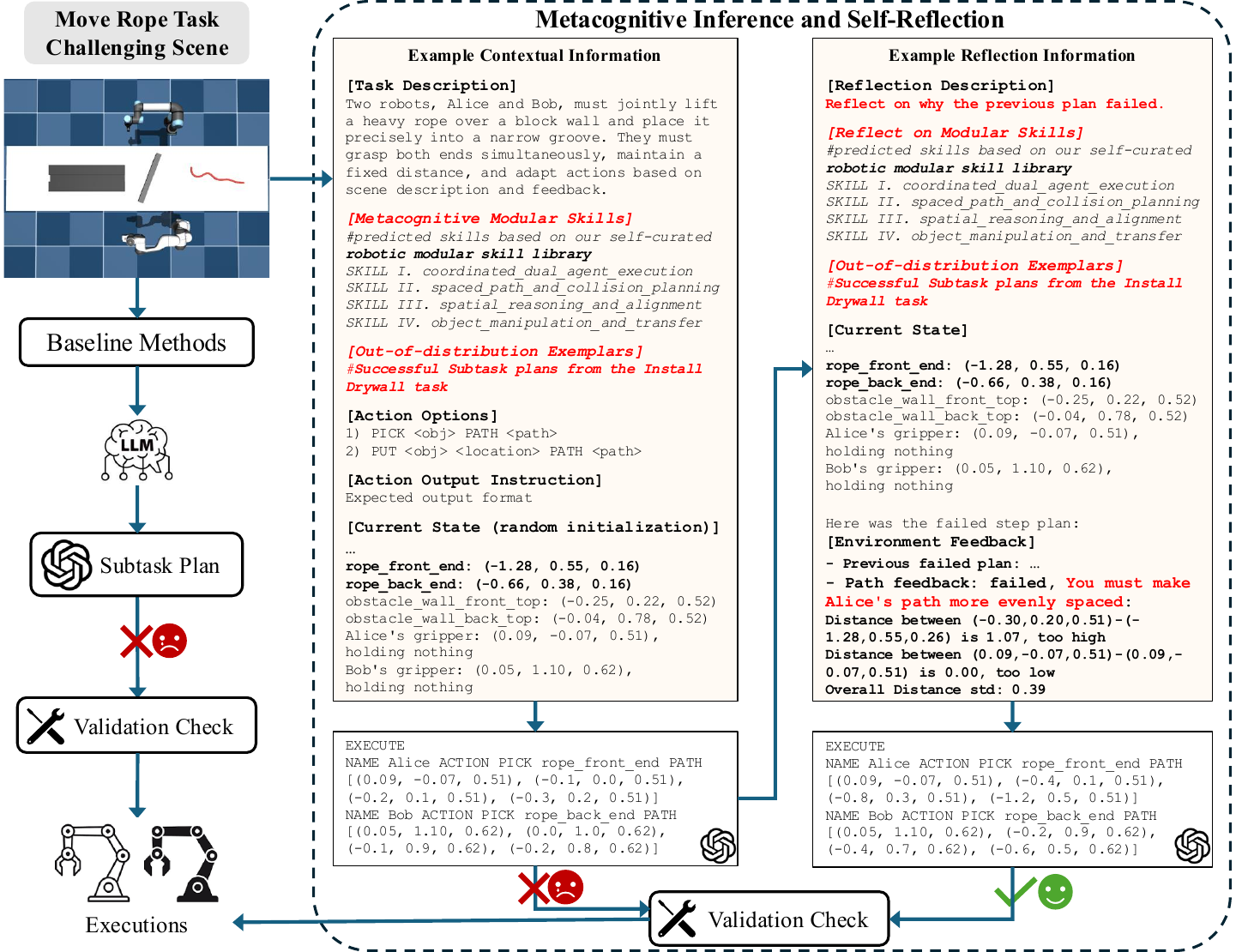}
  \caption{Demonstration of our metacognitive inference and self-reflection components. In most cases, the metacognitive inference component works well for generating valid subtask plans; if not, the self-reflection component can serve as a fix.}
  \label{fig:meta_reflection}
\end{figure*}

\subsection{Reliability Enhancement}The evaluation results comparing our REFLEX framework with the two baselines across the three tasks in RoCo are shown in Table~\ref{tab:rocobench_results}. In the \textit{Move Rope} task, which is the most challenging task in the experiments, our framework achieves a success rate of $0.76$, representing a $17\%$ improvement over RoCo+GPT-4 and a $26\%$ gain over Central Plan. It also reduces environment steps in successful runs to $2.0$, compared to $2.5$ for RoCo+GPT-4 and $2.3$ for Central Plan, and lowers replan attempts to $2.4$, compared to $3.1$ for RoCo+GPT-4 and $3.9$ for Central Plan. These results indicate more effective coordination and faster convergence under challenging spatial constraints. For the \textit{Arrange Cabinet} task, our framework achieves a success rate of $0.95$, outperforming RoCo by $20\%$ and the Central Plan by $5\%$. Additionally, it requires the same number of environment steps as the Central Plan while requiring fewer replan attempts than both baselines, suggesting improved planning robustness and reduced reliance on corrective execution. For the \textit{Make Sandwich} task, which involves long-horizon planning and strict stacking constraints, our framework achieves performance comparable to Central Plan. In comparison to the RoCo+GPT-4 baseline, it achieves a $15\%$ higher success rate and reduces the required environment steps, highlighting its ability to generalize to structurally complex tasks with minimal planning overhead, despite a slight increase in replan attempts. These results validate the effectiveness of our framework in advancing the capabilities of LLM-powered robot agents for completing complex tasks under zero-shot settings. They suggest that the proposed metacognitive learning module enables LLM-equipped robot agents to adaptively and proactively reason about and reflect on spatial, temporal, and structural challenges.

In the \textit{Install Drywall} task, our framework REFLEX also demonstrates substantial improvements in both task success rate and efficiency compared to the RoCo+GPT-4 baseline, which is shown in Table~\ref{tab:drywall_results}. Specifically, REFLEX utilizing LLaMA-3.1 achieves a task success rate of 0.95, representing a significant 53\% improvement over RoCo's success rate of 0.62. Additionally, it reduces both environment steps and replan attempts from 5.3 and 5.8 (baseline) to 4.2 and 1.9, respectively. When employing GPT-4, REFLEX achieves a high success rate of 1.00 with further reductions in environment steps, which is 4.1, and zero replan attempts. These results validate our framework's advanced effectiveness in handling complex coordination, real-time adaptability, and precise execution, thus significantly enhancing the reliability and efficiency of robotic task execution under challenging zero-shot conditions.

\subsection{Creativity Cultivation} In the experiments, we observe that our framework is capable of generating solutions that differ from the ground truth yet still successfully complete the tasks, which is demonstrated in Fig.~\ref{fig:creativity}. These findings support our hypothesis that metacognitive learning can cultivate structured creativity in robotic planning. To further illustrate this, we provide one representative case. In the \textit{Move Rope} task, two robot agents need to collaboratively grasp the ends of a rope, maneuver it over an obstacle wall, and place it into a designated groove. In the ground-truth plan, the robots grasp the two ends of the rope to complete the task. In contrast, after initial plan generation failures due to collision or IK infeasibility, our framework generates an alternative motion plan in which one robot grasps the rope slightly inward instead of at the end. Although this alternative plan deviates from the ground truth, it effectively shortens the trajectory, improves the spatial separation between the robot arms, and significantly reduces the risk of collisions and IK errors. 
\begin{table}[ht]
\caption{Reflection success rate on each task using our framework REFLEX.}
\centering
\begin{tabular}{lcc}
\toprule
Reflection Success Rate & LLaMA-3.1 & GPT-4 \\
\midrule
\textbf{Move Rope} & 0.44 & 0.50 \\
\textbf{Arrange Cabinet} & 1.00 & 0.91 \\
\textbf{Make Sandwich} & 0.70 & 1.00 \\
\textbf{Install Drywall} & 0.50 & 1.00 \\
\bottomrule
\end{tabular}
\label{tab:reflection_success}
\end{table}
\subsection{Self-Reflection Analysis} As shown in Fig.~\ref{fig:overview}, the self-reflection component is one of the three key elements of our proposed REFLEX. An example is shown in Fig.~\ref{fig:meta_reflection} to demonstrate the workflow of metacognitive inference and self-reflection modules. To evaluate its practical impact, we introduce an additional metric in our experiments, the \textit{Reflection Success Rate}, defined as the proportion of successful plan regenerations among all reflection attempts within task rounds that ultimately succeeded.

As shown in the Table~\ref{tab:reflection_success}, in the \textit{Arrange Cabinet} task, the framework achieves a perfect reflection success rate of $100\%$, indicating that every initial failure was successfully recovered through self-reflection. In the \textit{Make Sandwich} task, which involves long-horizon dependencies and stacking constraints, the framework recovers from $70\%$ of failed plans via self-reflection. Even in the most challenging \textit{Move Rope} task, characterized by tight spatial coordination, the framework achieves a $44\%$ reflection success rate, underscoring the role of self-reflection in enabling meaningful plan recovery under physical constraints.

In our newly introduced \textit{Install Drywall} task, the framework also demonstrates strong self-reflection capabilities. Specifically, using GPT-4 achieves an ideal reflection success rate of $100\%$, signifying perfect recovery from all initial failures through reflection-driven adjustments. With LLaMA-3.1, while slightly lower, our framework still achieves a notable reflection success rate of $50\%$, reinforcing the practical effectiveness of our metacognitive module in managing complex spatial coordination and real-time alignment corrections essential for this task.

These findings highlight the critical role of the self-reflection component within our REFLEX framework. They also show that the proposed framework not only supports LLM-powered agents in initial plan generation but also enables them to proactively reason about and reflect on execution failures, providing the essential capability of reliable recovery and adaptation in zero-shot settings.

\section{Discussions and Limitations}
\label{Discussion and Limitations}

Our experimental results demonstrate that our REFLEX framework significantly enhances the effectiveness and adaptability of LLM-powered robots in zero-shot planning. Across all tasks, including the newly proposed \textit{Install Drywall} benchmark, the integration of skill decomposition, metacognitive inference, and self-reflection leads to higher task success rates and fewer replans. Notably, in the \textit{Move Rope} task, our framework exhibits creative behavior by adapting grasp points to handle IK failures and avoid collisions. These adaptations, while deviating from ground-truth plans, remain valid and effective, highlighting the role of metacognition in enabling structured creativity. Our statistical analysis also confirms that our framework not only improves overall success but also facilitates graceful recovery from failure. Finally, we observe that even an open-source LLM, such as LLaMA-3.1, performs competitively with GPT-4 under our framework, which suggests that the gains stem from the structure of our method rather than the scale or proprietary nature of the foundation model.

While our initial exploration demonstrates strong zero-shot performance and emergent structured creativity, the findings are mainly based on empirical observations. A formal framework is needed to analyze how metacognitive learning enables creativity, and future work will incorporate ablation studies to isolate component contributions, which is currently challenging due to the framework’s early-stage nature.

\section{Conclusions and Future Work}
\label{Conclusion}

This paper presented REFLEX, a metacognitive learning framework that integrates modular skill abstraction, metacognitive inference, and self-reflection into LLM-powered multi-robot collaboration. Unlike prior prompt-based or purely reactive replanning approaches, REFLEX explicitly introduces a structured reasoning layer that enables robotic agents to diagnose failure causes, retrieve relevant modular skills, and iteratively synthesize improved motion plans under zero-shot conditions. Through comprehensive evaluation on RoCoBench and our newly proposed \textit{Install Drywall} benchmark, we demonstrate that REFLEX significantly improves task success rates, reduces replanning frequency, and enhances recovery from execution failures. Importantly, the framework enables the generation of operationally distinct yet valid solutions, highlighting the potential of structured metacognitive reasoning to support adaptive and robust robotic behavior. Importantly, beyond improving reliability, REFLEX enables the structured generation of alternative yet operationally valid motion plans that deviate from predefined ground-truth solutions. This behavior reflects the potential of metacognitive reasoning to support adaptive plan diversification under physical and operational constraints, highlighting a step toward more robust and flexible embodied intelligence.

Beyond empirical performance gains, our findings suggest that embedding metacognitive reasoning mechanisms into LLM-driven robotic systems provides a promising structural direction for advancing reliability, adaptability, and generalization in embodied AI. Future work will investigate formal analyses of metacognitive dynamics, scalable skill library expansion, and deployment in real-world multi-robot settings.

\bibliographystyle{IEEEtran}
\bibliography{iros_2026}

\end{document}